\documentclass[10pt,twocolumn,letterpaper]{article}

\usepackage{cvis}              % To produce the CAMERA-READY version
\definecolor{cvisblue}{rgb}{0.21,0.49,0.74}
\usepackage[pagebackref,breaklinks,colorlinks,allcolors=cvisblue]{hyperref}
\def\confName{CVIS}
\def\confYear{2026}

\title{Real-Time Physics Simulation with Dynamic Mesh-Gaussian Reconstructions}

\author{
Adrian Ramlal\\
University of Waterloo\\
{\tt\small adrian.ramlal@uwaterloo.ca}
\and
John S. Zelek\\
University of Waterloo\\
{\tt\small jzelek@uwaterloo.ca}
}

\begin{document}
\maketitle
% Abstract has a 150 word limit
\begin{abstract}
Integrating dynamic 3D reconstructions into physics simulation requires fixed mesh topology for efficient collision detection, but state-of-the-art methods like DG-Mesh produce varying topology optimized for geometric quality. We investigate whether topology conversion can enable physics integration while preserving reconstruction fidelity. We propose a dual-representation framework combining fixed-topology meshes for physics with Gaussian splatting for rendering, achieving 4.65$\times$ speedup over varying-topology baselines through runtime vertex buffer updates. We evaluate two conversion strategies, temporal correspondence tracking and template-based projection, against native fixed-topology methods (MaGS) on the DG-Mesh dataset. Our evaluation reveals that both conversion approaches incur 65--80\% geometric degradation, producing results inferior to MaGS despite DG-Mesh's superior initial quality. This demonstrates that high-quality reconstruction and physics-compatible topology represent fundamentally distinct objectives that cannot be reconciled through post-processing. Our findings inform future development of physics-aware reconstruction methods and our framework enables real-time simulation with any fixed-topology approach.
\end{abstract}    
\section{Introduction}
\label{sec:intro}

Physics simulation with photorealistic rendering is essential for robotics~\cite{todorov2012mujoco}, virtual reality~\cite{pechko2025gsversemeshbasedgaussiansplatting}, and reinforcement learning~\cite{genesis2024}. Training robots in simulation requires realistic collision dynamics and visual feedback to enable effective sim-to-real transfer. Similarly, embodied AI agents need physically accurate environments to develop transferable behaviors. Recent advances in neural rendering, particularly 3D Gaussian Splatting (3DGS)~\cite{kerbl3Dgaussians} and its dynamic extensions~\cite{liu2024dynamic,mags2024}, enable photorealistic reconstruction of dynamic scenes from multi-view video with explicit mesh geometry. However, integrating them into physics engines faces a critical barrier: \emph{mesh topology consistency}.

Modern physics engines require \emph{fixed topology} (consistent vertex count and face connectivity) to maintain efficient collision detection structures. Varying topology necessitates expensive per-frame rebuilds of spatial acceleration structures, preventing real-time simulation.

This exposes a fundamental trade-off in current methods. DG-Mesh~\cite{liu2024dynamic} achieves state-of-the-art quality via iso-surface extraction but produces \emph{varying topology}. MaGS~\cite{mags2024} maintains \emph{fixed topology} by jointly optimizing a template mesh with Gaussians, enabling physics simulation at the cost of fidelity. Can high-quality varying-topology reconstructions be converted to fixed topology while preserving geometric quality? Bridging this gap would enable leveraging the highest-quality reconstructions for physics simulation in robotics and interactive applications.

We investigate this question through systematic evaluation in the Genesis simulator~\cite{genesis2024}. Our contributions are:

\begin{enumerate}
    \item \textbf{Dual-Representation Framework.} We develop an architecture combining fixed-topology meshes for physics with Gaussians for rendering. Runtime vertex buffer updates achieve 147.8 FPS versus 31.8 FPS for varying topology (4.65× speedup), enabling efficient integration of any fixed-topology method.
    
    \item \textbf{Topology Conversion Evaluation.} We propose two conversion strategies (temporal tracking and template projection) and provide the first systematic comparison against native fixed-topology methods on the DG-Mesh dataset~\cite{liu2024dynamic}, measuring geometric accuracy across six dynamic scenes.
\end{enumerate}

Our experiments reveal that topology conversion incurs 65--80\% geometric degradation, making both strategies inferior to native fixed-topology methods. This demonstrates that high-quality reconstruction and physics-compatible topology are fundamentally distinct objectives. Our framework provides valuable infrastructure for fixed-topology methods and our evaluation informs future physics-aware reconstruction approaches.
\section{Related Works}
\label{sec:related_work}

% DG-Mesh ~\cite{liu2024dynamic}
% MaGS ~\cite{mags2024}
% 3DGS ~\cite{kerbl3Dgaussians}
% Genesis ~\cite{genesis2024}

\subsection{Dynamic Neural Rendering}
Neural Radiance Fields (NeRF)~\cite{mildenhall2020nerf} and extensions~\cite{pumarola2020d,park2021nerfies,gao2021dynamic} model dynamic scenes with volumetric representations but lack explicit geometry. 3D Gaussian Splatting (3DGS)~\cite{kerbl3Dgaussians} provides explicit point-based rendering with real-time performance. Dynamic 3DGS methods~\cite{wu20234dgs,yang2023deformable3dgs,luiten2023dynamic} learn deformation networks to model temporal motion, with controllable variants like SC-GS~\cite{huang2023scgs} using sparse control points for editing. While achieving high rendering quality, these point-based methods produce meshes with inconsistent topology across frames, preventing direct physics integration.

\subsection{Mesh-Gaussian Hybrid Representations}
Recent work combines meshes with 3DGS for structured geometry. SuGaR~\cite{guedon2023sugar} regularizes Gaussians to surfaces and extracts meshes via Poisson reconstruction. GaMeS~\cite{waczynska2024games} introduces mesh-Gaussian binding for topology-preserving edits through triangle soups. Domain-specific methods leverage parametric models: GaussianAvatars~\cite{qian2024gaussianavatars} uses FLAME for animatable heads, while PGC~\cite{guo2025pgc} and Gaussian Garments~\cite{rong2024gaussiangarments} bind Gaussians to cloth meshes for physics-based simulation.

For general dynamic scenes, DG-Mesh~\cite{liu2024dynamic} achieves state-of-the-art geometric quality through differentiable Poisson reconstruction and Marching Cubes, with Gaussian-Mesh Anchoring enforcing one-to-one Gaussian-face correspondence. However, the iso-surface extraction produces varying topology across frames. In contrast, MaGS~\cite{mags2024} maintains fixed topology by jointly optimizing a template mesh with mesh-adsorbed Gaussians, incorporating deformation priors like ARAP~\cite{ARAP_modeling:2007} and SMPL~\cite{SMPL:2015}. This prioritizes topological stability and physics compatibility over geometric fidelity. Our work provides the first systematic comparison of these approaches for physics simulation.

\subsection{Physics Integration with Neural Representations}
Physics engines require fixed topology for efficient collision detection through spatial acceleration structures (BVH trees, SDFs). Traditional approaches use explicit meshes with simulators like MuJoCo~\cite{todorov2012mujoco}, Bullet~\cite{coumans2021bullet}, or Genesis~\cite{genesis2024}. Recent neural-physics integration includes PhysGaussian~\cite{xie2023physgaussian} with mass-spring dynamics on Gaussian particles, GASP~\cite{borycki2024gasp} for soft-body simulation on point clouds, and Splatting Physical Scenes~\cite{moran2025splattingphysicalscenesendtoend} optimizing coarse meshes through differentiable simulation. However, these works either avoid mesh extraction or use simplified static geometry, leaving a gap in integrating high-fidelity time-varying meshes.

Our framework enables systematic evaluation of topology strategies for physics simulation in Genesis~\cite{genesis2024}. We investigate whether topology conversion can preserve geometric quality while enabling physics integration, and provide infrastructure for real-time simulation with fixed-topology methods through a dual-representation architecture that decouples physics meshes from photorealistic rendering.

\section{Method}
\label{sec:method}

Our method enables integration of state-of-the-art dynamic reconstruction techniques into real-time physics simulation through a dual-representation framework that decouples physics-compatible mesh topology from high-fidelity rendering.

\subsection{Background: Dynamic Reconstruction}

We build upon two complementary approaches representing different points in the quality-topology trade-off space.

\textbf{DG-Mesh}~\cite{liu2024dynamic} achieves state-of-the-art reconstruction quality (CD = 0.697) by representing scenes using canonical 3D Gaussians deformed via learned temporal networks. Geometry extraction via Differentiable Poisson Surface Reconstruction and Marching Cubes produces high-quality meshes with \emph{varying topology}: both vertex count $|\mathbf{V}_t|$ and connectivity $\mathbf{F}_t$ change across frames, preventing direct physics integration.

\textbf{MaGS}~\cite{mags2024} maintains \emph{fixed topology} by jointly optimizing a template mesh and Gaussian primitives. Gaussians are adsorbed to mesh faces and displaced via learned networks, while mesh vertices deform through a deformation network. The fixed template topology enables direct physics integration but achieves lower geometric fidelity (CD = 1.108) than DG-Mesh.

DG-Mesh's superior geometric quality (36\% lower CD) motivates our investigation of topology conversion strategies to enable physics integration while preserving reconstruction fidelity.

\subsection{Physics-Rendering Dual Representation}

Our framework maintains two synchronized representations: a \textbf{physics mesh} ($\mathcal{M}_{\text{phys}}$) with fixed topology for collision detection, and a \textbf{rendering representation} ($\mathcal{R}_{\text{render}}$) using high-fidelity Gaussian splatting for photorealistic visualization. This decoupling enables optimal performance for both physics and rendering while supporting comparative analysis of different reconstruction methods.

\subsection{Fixed-Topology Mesh Integration}

\paragraph{MaGS: Native Fixed-Topology.}
MaGS meshes directly provide fixed topology. At simulation time $t$, we load corresponding deformed vertices: $k = \lfloor t \cdot f_{\text{recon}} \rfloor$ where $f_{\text{recon}}$ is the reconstruction framerate.

\paragraph{DG-Mesh: Topology Conversion Strategies.}
To enable physics integration of DG-Mesh's varying-topology sequences $\{\mathcal{M}_t = (\mathbf{V}_t, \mathbf{F}_t)\}$, we explore two conversion strategies to establish fixed correspondences.

\textbf{Strategy 1: Temporal Correspondence Tracking.}
Given an initial template $\mathcal{M}_0 = (\mathbf{V}_0, \mathbf{F}_0)$, we track vertices forward through time:
\begin{equation}
\tilde{\mathbf{V}}_{t+1} = \{\text{NearestSurface}(\mathbf{v}_i^t, \mathcal{M}_{t+1})\}_{i=1}^{N_v}
\end{equation}
where $\text{NearestSurface}(\mathbf{v}, \mathcal{M})$ projects point $\mathbf{v}$ onto mesh $\mathcal{M}$ via k-NN weighted surface projection. To mitigate drift accumulation, we periodically re-anchor to the template every 50 frames by blending 80\% tracked motion with 20\% direct template projection.

\textbf{Strategy 2: Template-Based Projection.}
We independently project a static template onto each frame:
\begin{equation}
\tilde{\mathbf{V}}_t = \{\text{NearestSurface}(\mathbf{v}_i^{(0)}, \mathcal{M}_t)\}_{i=1}^{N_v}
\end{equation}
using barycentric interpolation for sub-face accuracy. This eliminates drift but may lose detail in regions with varying geometric complexity.

\subsection{Efficient Runtime Updates}

Our key technical contribution enables time-varying mesh geometry without rebuilding collision structures.

\textbf{Initialization.} We construct spatial acceleration structures (BVH, SDF) from template topology $\mathbf{F}_0$. These structures store only vertex and face \emph{indices}, enabling reuse across all timesteps.

\textbf{Per-Frame Update.} At each simulation step:
\begin{enumerate}[leftmargin=*]
\item Compute frame index $k = \lfloor t \cdot f_{\text{recon}} \rfloor$ and load vertices $\mathbf{V}_k$
\item Update vertex buffer via $O(N_v)$ memory copy
\item Recompute per-face normals on-demand during collision queries
\end{enumerate}

This approach scales as $O(N_v)$ per frame versus $O(N_v \log N_v)$ for BVH reconstruction, enabling real-time rates for typical meshes ($N_v \approx 10^3$--$10^4$).

\section{Experiments}
\label{sec:experiments}

We evaluate our framework on the DG-Mesh dataset~\cite{liu2024dynamic}, which provides ground truth mesh sequences for six dynamic scenes. Each sequence contains 200 frames with varying-topology meshes extracted from multi-view captures.

\subsection{Baseline: Varying-Topology Meshes}

To establish the computational necessity of fixed topology, we implement a baseline that uses DG-Mesh's original varying-topology meshes directly in physics simulation. At each frame, this baseline must destroy and rebuild all collision structures (BVH trees, contact caches) to accommodate topology changes.

We measure performance on a test scene containing the \textit{horse} sequence with 5 rigid spheres falling onto the deforming mesh. The varying-topology baseline achieves 31.8 FPS, with collision rebuild overhead consuming 23.8 ms per frame (75.6\% of total frame time). While this provides the highest geometric quality (CD = 0.697), the performance is insufficient for real-time applications. In contrast, fixed-topology approaches achieve 147.8 FPS by eliminating rebuild overhead through vertex buffer updates, demonstrating a 4.65× speedup.

\subsection{Geometric Accuracy Analysis}

We compare four approaches: DG-Mesh original (varying topology), our two topology conversion strategies (temporal tracking and template projection), and MaGS native (fixed topology). Like DG-Mesh~\cite{liu2024dynamic} and MaGS~\cite{mags2024}, we measure quality using Chamfer Distance (CD, in $10^{-3}$ units) and Earth Mover's Distance (EMD, in $10^{-1}$ units) by sampling 10K points from reconstructed and ground truth meshes.

\begin{table}[t]
\centering
\caption{Geometric reconstruction quality on DG-Mesh dataset averaged over 6 scenes. Lower is better. Our topology conversion strategies result in significant degradation compared to both original DG-Mesh and MaGS native output.}
\label{tab:geometric_accuracy}
\begin{tabular}{lcc}
\toprule
\textbf{Method} & \textbf{CD} $\downarrow$ & \textbf{EMD} $\downarrow$ \\
\midrule
DG-Mesh (Original) & \textbf{0.697} & 0.130 \\
\midrule
Temporal Tracking (Ours) & 1.152 & 0.185 \\
Template Projection (Ours) & 1.254 & 0.200 \\
\midrule
MaGS (Native) & 1.108 & \textbf{0.113} \\
\bottomrule
\end{tabular}
\vspace{-0.5em}
\end{table}

Table~\ref{tab:geometric_accuracy} shows that DG-Mesh's original varying-topology meshes achieve the best accuracy (CD = 0.697, EMD = 0.130). MaGS achieves CD = 1.108 and EMD = 0.113. While MaGS has higher CD, it achieves lower EMD, indicating better global shape preservation consistent with its fixed-topology design.

Our topology conversion strategies result in substantial degradation. Temporal tracking achieves CD = 1.152 and EMD = 0.185 (65\% and 42\% worse than original DG-Mesh). Template projection performs worse with CD = 1.254 and EMD = 0.200 (80\% and 54\% degradation). 

Critically, both conversion strategies produce geometrically inferior results compared to MaGS native output. Temporal tracking achieves 4\% worse CD and 64\% worse EMD than MaGS, while template projection is 13\% worse in CD and 77\% worse in EMD. This demonstrates a fundamental limitation: converting high-quality varying-topology reconstruction to fixed-topology incurs geometric losses that negate the original quality advantage.

\subsection{Qualitative Analysis}

\begin{figure}[t]
\centering
\includegraphics[width=\linewidth]{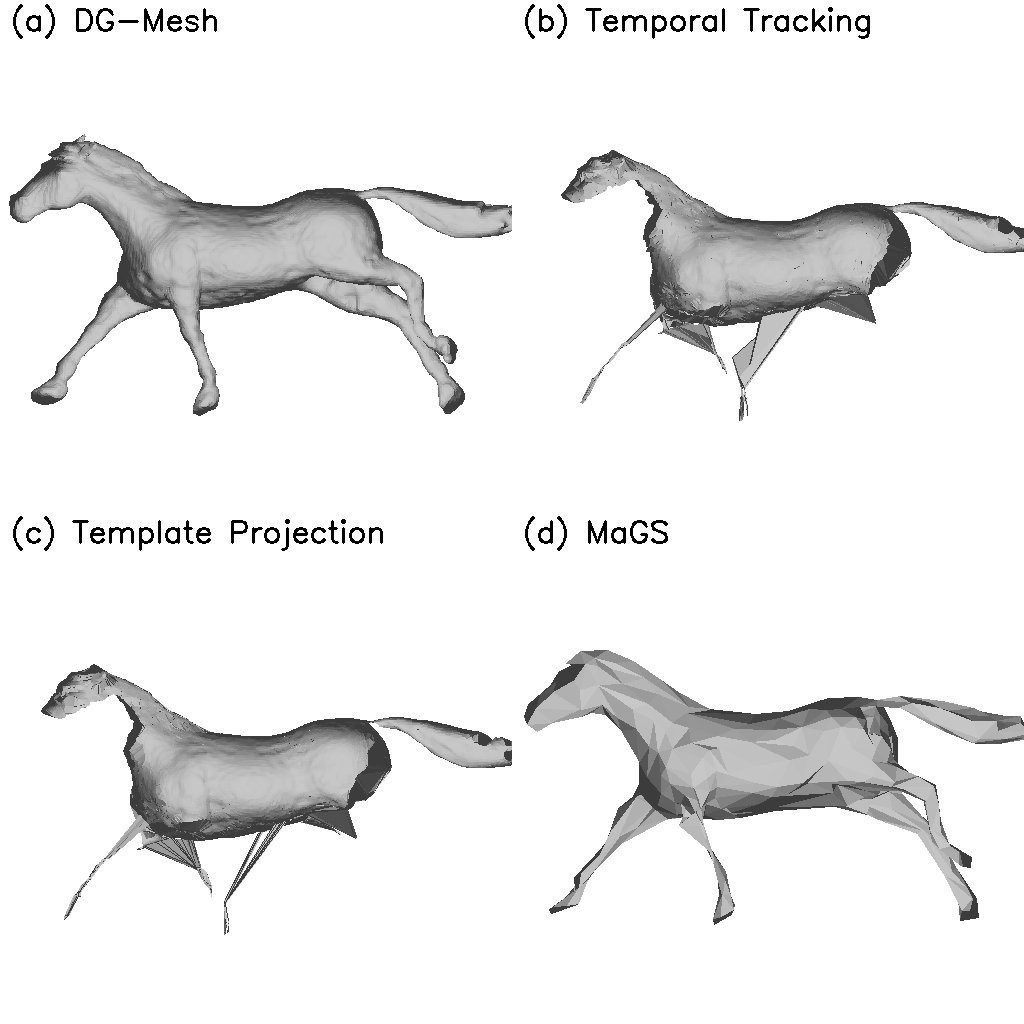}
\caption{Qualitative comparison at frame 20 on the \textit{horse} sequence. From left to right: (a) DG-Mesh original varying-topology mesh, (b) our temporal tracking conversion, (c) our template projection conversion, (d) MaGS native fixed-topology. Both conversion strategies introduce visible artifacts compared to the original DG-Mesh quality. MaGS maintains better geometric fidelity despite having native fixed topology.}
\label{fig:qualitative_comparison}
\end{figure}

Figure~\ref{fig:qualitative_comparison} visualizes geometric quality at frame 20 of the \textit{horse} sequence. The original DG-Mesh mesh exhibits crisp surface details and accurate thin structures. Our temporal tracking conversion introduces noticeable smoothing and vertex drift, particularly visible in the neck. Template projection shows more severe degradation with over-smoothed surfaces and loss of fine features. In comparison, MaGS native output maintains substantially better geometric fidelity than either conversion strategy.

The temporal tracking method suffers from accumulated drift over the 200-frame sequence despite periodic re-anchoring, while template projection loses detail by forcing a fixed vertex distribution onto varying geometric complexity. These results reveal that the topology conversion process fundamentally compromises the geometric quality that made DG-Mesh attractive for this application.

\section{Discussion}
\label{sec:discussion}

Our experiments reveal a critical finding: converting varying-topology reconstructions to fixed topology for physics simulation fundamentally degrades geometric quality (65--80\% increase in CD, 42--54\% increase in EMD). Both our conversion strategies incur substantial geometric losses that eliminate DG-Mesh's quality advantage and produce results inferior to MaGS's native fixed-topology method.

This finding demonstrates that high-quality reconstruction and physics-compatible topology represent \emph{distinct design objectives} that cannot be easily bridged through post-processing. DG-Mesh optimizes for geometric fidelity without topological constraints, while MaGS balances quality with topological stability from the outset. The underlying cause is that topology conversion requires establishing correspondences between meshes with different geometric complexity distributions. Temporal tracking accumulates drift as errors compound over frames, while template projection forces a fixed vertex distribution onto varying geometric features.

Despite these conversion limitations, our work provides valuable contributions. The dual-representation architecture and runtime vertex buffer update mechanism achieve a 4.65× performance improvement, enabling real-time physics simulation at 147.8 FPS. This infrastructure supports any fixed-topology dynamic reconstruction method. Our systematic evaluation provides the first direct comparison between varying-topology conversion strategies and native fixed-topology approaches in physics simulation contexts, offering concrete evidence that informs future research directions.

\textbf{Limitations.} Our topology conversion strategies assume smooth deformations without topological events (self-contact, tears). Temporal tracking accumulates drift beyond 200 frames, while template projection cannot adapt vertex density to local complexity. Our evaluation focuses on geometric accuracy (CD, EMD) rather than physics-specific metrics like collision fidelity or task performance. The framework is tested only with rigid body physics; soft-body dynamics and deformable materials remain unexplored. Fundamentally, fixed-topology representations cannot capture phenomena requiring topology changes.

\section{Conclusion}
\label{sec:conclusion}

We identified mesh topology consistency as the critical barrier preventing integration of state-of-the-art dynamic 3D reconstructions into real-time physics simulation. We demonstrated that varying-topology meshes impose a 4.65× performance penalty due to per-frame collision structure rebuilds. We explored two topology conversion strategies to enable physics integration of high-quality reconstructions, revealing that both approaches incur 65--80\% geometric degradation and produce results inferior to native fixed-topology methods like MaGS. This demonstrates that high-quality reconstruction and physics-compatible topology are fundamentally distinct objectives that cannot be easily reconciled through post-processing. Our dual-representation framework provides valuable infrastructure for integrating fixed-topology methods into physics simulation, achieving 147.8 FPS with runtime vertex buffer updates. Future work should focus on developing reconstruction methods that directly optimize for high-fidelity, fixed-topology constraints during training rather than attempting post-hoc conversion.

{
    \small
    \bibliographystyle{ieeenat_fullname}
    \bibliography{main}

@String(CVPR= {IEEE Conf. Comput. Vis. Pattern Recog.})

@String(ICCV= {Int. Conf. Comput. Vis.})

@String(ECCV= {Eur. Conf. Comput. Vis.})

@String(CVPR  = {CVPR})

@String(ICCV  = {ICCV})

@String(ECCV  = {ECCV})

@misc{liu2024dynamic,
 title={Dynamic Gaussians Mesh: Consistent Mesh Reconstruction from Monocular Videos}, 
 author={Isabella Liu and Hao Su and Xiaolong Wang},
 year={2024},
 eprint={2404.12379},
 archivePrefix={arXiv},
 primaryClass={cs.CV}
}

@article{mags2024,
  title={MaGS: Reconstructing and Simulating Dynamic 3D Objects with Mesh-adsorbed Gaussian Splatting},
  author={Shaojie Ma and Yawei Luo and Wei Yang and Yi Yang},
  year={2024},
  eprint={2406.01593},
  archivePrefix={arXiv},
  primaryClass={cs.CV},
  url={https://arxiv.org/abs/2406.01593}
}

@misc{genesis2024,
  author = {Genesis Authors},
  title = {Genesis: A Generative and Universal Physics Engine for Robotics and Beyond},
  month = {December},
  year = {2024},
  url = {https://github.com/Genesis-Embodied-AI/Genesis}
}

@Article{kerbl3Dgaussians,
      author       = {Kerbl, Bernhard and Kopanas, Georgios and Leimk{\"u}hler, Thomas and Drettakis, George},
      title        = {3D Gaussian Splatting for Real-Time Radiance Field Rendering},
      journal      = {ACM Transactions on Graphics},
      number       = {4},
      volume       = {42},
      month        = {July},
      year         = {2023},
      url          = {https://repo-sam.inria.fr/fungraph/3d-gaussian-splatting/}
}

@article{pumarola2020d,
  title={D-NeRF: Neural Radiance Fields for Dynamic Scenes},
  author={Pumarola, Albert and Corona, Enric and Pons-Moll, Gerard and Moreno-Noguer, Francesc},
  journal={arXiv preprint arXiv:2011.13961},
  year={2020}
}

@inproceedings{mildenhall2020nerf,
 title={NeRF: Representing Scenes as Neural Radiance Fields for View Synthesis},
 author={Ben Mildenhall and Pratul P. Srinivasan and Matthew Tancik and Jonathan T. Barron and Ravi Ramamoorthi and Ren Ng},
 year={2020},
 booktitle={ECCV},
}

@article{park2021nerfies,
  author    = {Park, Keunhong and Sinha, Utkarsh and Barron, Jonathan T. and Bouaziz, Sofien and Goldman, Dan B and Seitz, Steven M. and Martin-Brualla, Ricardo},
  title     = {Nerfies: Deformable Neural Radiance Fields},
  journal   = {ICCV},
  year      = {2021},
}

@InProceedings{wu20234dgs,
    author    = {Wu, Guanjun and Yi, Taoran and Fang, Jiemin and Xie, Lingxi and Zhang, Xiaopeng and Wei, Wei and Liu, Wenyu and Tian, Qi and Wang, Xinggang},
    title     = {4D Gaussian Splatting for Real-Time Dynamic Scene Rendering},
    booktitle = {Proceedings of the IEEE/CVF Conference on Computer Vision and Pattern Recognition (CVPR)},
    month     = {June},
    year      = {2024},
    pages     = {20310-20320}
}

@inproceedings{luiten2023dynamic,
  title={Dynamic 3D Gaussians: Tracking by Persistent Dynamic View Synthesis},
  author={Luiten, Jonathon and Kopanas, Georgios and Leibe, Bastian and Ramanan, Deva},
  booktitle={3DV},
  year={2024}
}

@Article{waczynska2024games,
      author         = {Joanna Waczyńska and Piotr Borycki and Sławomir Tadeja and Jacek Tabor and Przemysław Spurek},
      title          = {GaMeS: Mesh-Based Adapting and Modification of Gaussian Splatting},
      year           = {2024},
      eprint         = {2402.01459},
      archivePrefix  = {arXiv},
      primaryClass   = {cs.CV},
}

@article{guedon2023sugar,
title={SuGaR: Surface-Aligned Gaussian Splatting for Efficient 3D Mesh Reconstruction and High-Quality Mesh Rendering},
author={Gu{\'e}don, Antoine and Lepetit, Vincent},
journal={CVPR},
year={2024}
}

@inproceedings{qian2024gaussianavatars,
  title={Gaussianavatars: Photorealistic head avatars with rigged 3d gaussians},
  author={Qian, Shenhan and Kirschstein, Tobias and Schoneveld, Liam and Davoli, Davide and Giebenhain, Simon and Nie{\ss}ner, Matthias},
  booktitle={Proceedings of the IEEE/CVF Conference on Computer Vision and Pattern Recognition},
  pages={20299--20309},
  year={2024}
}

@INPROCEEDINGS{guo2025pgc,
  author={Guo, Michelle and Chiang, Matt Jen-Yuan and Santesteban, Igor and Sarafianos, Nikolaos and Chen, Hsiao-Yu and Halimi, Oshri and Božič, Aljaž and Saito, Shunsuke and Wu, Jiajun and Liu, C. Karen and Stuyck, Tuur and Larionov, Egor},
  booktitle={2025 IEEE/CVF Conference on Computer Vision and Pattern Recognition (CVPR)}, 
  title={PGC: Physics-Based Gaussian Cloth from a Single Pose}, 
  year={2025},
  volume={},
  number={},
  pages={21215-21225},
  doi={10.1109/CVPR52734.2025.01976}}

@inproceedings{rong2024gaussiangarments,
  title={{Gaussian Garments}: Reconstructing Simulation-Ready Clothing with Photorealistic Appearance from Multi-View Video}, 
  author={Boxiang Rong and Artur Grigorev and Wenbo Wang and Michael J. Black and Bernhard Thomaszewski and Christina Tsalicoglou and Otmar Hilliges},
  booktitle={International Conference on 3D Vision 2025},
  year={2025}
}

@article{huang2023scgs,
  title={SC-GS: Sparse-Controlled Gaussian Splatting for Editable Dynamic Scenes},
  author={Huang, Yi-Hua and Sun, Yang-Tian and Yang, Ziyi and Lyu, Xiaoyang and Cao, Yan-Pei and Qi, Xiaojuan},
  journal={arXiv preprint arXiv:2312.14937},
  year={2023}
}

@inproceedings{ARAP_modeling:2007,
    author = {Olga Sorkine and Marc Alexa},
    title = {As-Rigid-As-Possible Surface Modeling},
    booktitle = {Proceedings of EUROGRAPHICS/ACM SIGGRAPH Symposium on Geometry Processing},
    year = {2007},
    pages = {109--116},
}

@article{SMPL:2015,
  author = {Loper, Matthew and Mahmood, Naureen and Romero, Javier and Pons-Moll, Gerard and Black, Michael J.},
  title = {{SMPL}: A Skinned Multi-Person Linear Model},
  journal = {ACM Trans. Graphics (Proc. SIGGRAPH Asia)},
  month = oct,
  number = {6},
  pages = {248:1--248:16},
  publisher = {ACM},
  volume = {34},
  year = {2015}
}

@misc{moran2025splattingphysicalscenesendtoend,
      title={Splatting Physical Scenes: End-to-End Real-to-Sim from Imperfect Robot Data}, 
      author={Ben Moran and Mauro Comi and Arunkumar Byravan and Steven Bohez and Tom Erez and Zhibin Li and Leonard Hasenclever},
      year={2025},
      eprint={2506.04120},
      archivePrefix={arXiv},
      primaryClass={cs.RO},
      url={https://arxiv.org/abs/2506.04120}, 
}

@Article{borycki2024gasp,
      author={Piotr Borycki and Weronika Smolak and Joanna Waczyńska and Marcin Mazur and Sławomir Tadeja and Przemysław Spurek},
      title={GASP: Gaussian Splatting for Physic-Based Simulations},
      year={2024},
      eprint={2409.05819},
      archivePrefix={arXiv},
      primaryClass={cs.CV},
      url={https://arxiv.org/abs/2409.05819}, 
}

@article{xie2023physgaussian,
      title={PhysGaussian: Physics-Integrated 3D Gaussians for Generative Dynamics}, 
      author={Xie, Tianyi and Zong, Zeshun and Qiu, Yuxing and Li, Xuan and Feng, Yutao and Yang, Yin and Jiang, Chenfanfu},
      journal={arXiv preprint arXiv:2311.12198},
      year={2023},
}

@inproceedings{todorov2012mujoco,
  title={MuJoCo: A physics engine for model-based control},
  author={Todorov, Emanuel and Erez, Tom and Tassa, Yuval},
  booktitle={2012 IEEE/RSJ International Conference on Intelligent Robots and Systems},
  pages={5026--5033},
  year={2012},
  organization={IEEE},
  doi={10.1109/IROS.2012.6386109}
}

@MISC{coumans2021bullet,
author =   {Erwin Coumans and Yunfei Bai},
title =    {PyBullet, a Python module for physics simulation for games, robotics and machine learning},
howpublished = {\url{http://pybullet.org}},
year = {2016--2021}
}

@inproceedings{gao2021dynamic,
  Author    = {Gao, Chen and Saraf, Ayush and Kopf, Johannes and Huang, Jia-Bin},
  Title     = {Dynamic View Synthesis from Dynamic Monocular Video},
  booktitle = {Proceedings of the IEEE International Conference on Computer Vision},
  year      = {2021}
}

@article{yang2023deformable3dgs,
    title={Deformable 3D Gaussians for High-Fidelity Monocular Dynamic Scene Reconstruction},
    author={Yang, Ziyi and Gao, Xinyu and Zhou, Wen and Jiao, Shaohui and Zhang, Yuqing and Jin, Xiaogang},
    journal={arXiv preprint arXiv:2309.13101},
    year={2023}
}

@misc{pechko2025gsversemeshbasedgaussiansplatting,
      title={GS-Verse: Mesh-based Gaussian Splatting for Physics-aware Interaction in Virtual Reality}, 
      author={Anastasiya Pechko and Piotr Borycki and Joanna Waczyńska and Daniel Barczyk and Agata Szymańska and Sławomir Tadeja and Przemysław Spurek},
      year={2025},
      eprint={2510.11878},
      archivePrefix={arXiv},
      primaryClass={cs.GR},
      url={https://arxiv.org/abs/2510.11878}, 
}
}

% WARNING: do not forget to delete the supplementary pages from your submission 
% \input{sec/X_suppl}

\end{document}